%% file: arxiv.tex
\documentclass{article} 

\input{math_commands.tex}

\usepackage{hyperref}
\usepackage{url}
\usepackage{xcolor}
\usepackage[normalem]{ulem}

\title{The Importance of the Current Input in\\ Sequence Modeling}

\author{Christian Oliva \& Luis F. Lago-Fernández\\
Departamento de Ingeniería Informática\\
Universidad Autónoma de Madrid\\
Madrid, 28049, Spain \\
\texttt{christian.oliva@estudiante.uam.es,luis.lago@uam.es}}

\begin{document}

\maketitle

\begin{abstract}
The last advances in sequence modeling are mainly based on deep learning approaches. The current state of the art involves the use of variations of the standard LSTM architecture, combined with several tricks that improve the final prediction rates of the trained neural networks. However, in some cases, these adaptations might be too much tuned to the particular problems being addressed. In this article, we show that a very simple idea, to add a direct connection between the input and the output, skipping the recurrent module, leads to an increase of the prediction accuracy in sequence modeling problems related to natural language processing. Experiments carried out on different problems show that the addition of this kind of connection to a recurrent network always improves the results, regardless of the architecture and training-specific details. When this idea is introduced into the models that lead the field, the resulting networks achieve a new state-of-the-art perplexity in language modeling problems.
\end{abstract}

\section{Introduction}

Deep learning models constitute the current state of the art in most artificial intelligence applications, from computer vision to robotics or medicine. When dealing with sequential data, Recurrent Neural Networks (RNNs), specially those architectures with gating mechanisms such as the LSTM \cite{DBLP:journals/neco/HochreiterS97}, the GRU \cite{DBLP:conf/emnlp/ChoMGBBSB14} and other variants, are usually the default choice. One of the most interesting applications of RNNs is related to the field of Natural Language Processing, where most tasks, such as machine translation, document summarization or language modeling, involve the manipulation of sequences of textual data. Of these, language modeling has been extensively used to test different innovations in recurrent architectures, mainly due to the ease of obtaining very large datasets that can be used to train neural networks with millions of parameters. 

Sequence modeling consists of predicting the next element in a sequence given the past history. In language modeling, the sequence is a text, and hence the task is to predict the next word or the next character. In this context, some of the best performing architectures include the Mogrifier LSTM \cite{mogrifier} and different variations of the AWD-LSTM \cite{awd_lstm}, usually combined with dynamic evaluation and mixture of sofmaxes (MoS) \cite{pmlr-v97-wang19f,frage}. These models obtain the best state-of-the-art performance with moderate size datasets, such as the Penn Treebank \cite{dyneval_Mikolov10} or the Wikitext-2 \cite{merity_wikitext2} corpora, when no additional data are used during training. When larger datasets are considered, or when external data are used to pre-train the networks, attention-based architectures usually outperform other models \cite{Radford2019LanguageMA,NEURIPS2020_1457c0d6}. 

In this work we use moderate-scale language modeling datasets to explore the effect of a mechanism recently proposed by \cite{Oliva21_DualRNN}, when combined with different LSTM-based models in the language modeling context. The idea consists of modifying a recurrent architecture by introducing a direct connection between the input and the output of the recurrent module. This has been shown to improve both the model's generalization results and its readability in simple tasks related to the recognition of regular languages. 

In a standard RNN, the output depends only on the network's hidden state, ${\displaystyle \vh}_{t}$, which in turn depends on both the input, ${\displaystyle \vx}_{t}$, and the recent past, ${\displaystyle \vh}_{t-1}$. But there is no explicit dependence of the network's output on its input. In some cases this could be a shortcoming, since the transformation of ${\displaystyle \vx}_{t}$ needed to compute the network's internal state is not necessarily the most appropriate to compute the output. However, an explicit dependence of the output on ${\displaystyle \vx}_{t}$ can be forced by adding a \textit{dual} connection that skips the recurrent layers. We claim that this strategy may be of general application in RNN models. 

To test our hypothesis we perform a thorough comparison of several state-of-the-art RNN architectures, with and without the \textit{dual} connection, on the Penn Treebank and the Wikitext-2 datasets. Our results show that, under all experimental conditions, the \textit{dual} architectures outperform their non-dual counterparts. In addition, the Mogrifier-LSTM enhanced with a \textit{dual} connection establishes a new state-of-the-art word-level  perplexity for the Penn Treebank problem when no additional data are used to train the models. 

The remainder of the article is organized as follows. First, in section \ref{sec:models}, we present the different models we have used and the two possible architectures, the standard recurrent architecture and the \textit{dual} architecture. In section \ref{sec:experiments}, we describe the datasets and the experimental setup. In section \ref{sec:results}, we present our results. And finally, in section \ref{sec:conclusion}, we extract some conclusions and discuss further lines of research.

\section{Models}
\label{sec:models}

We start by presenting the standard recurrent architecture which is common to all the models. In absence of a \textit{dual} connection, the basic architecture involves an embedding layer, a recurrent layer and a fully-connected layer with $softmax$ activation:

\begin{eqnarray}
    {\displaystyle \ve}_t &=& {\displaystyle \mW}^{ex} {\displaystyle \vx}_t \label{eq:embedding}\\
    {\displaystyle \vh}_t &=& REC({\displaystyle \ve}_t, {\displaystyle \sS}_{t-1}) \label{eq:rec}\\
    {\displaystyle \vy}_t &=& softmax({\displaystyle \mW}^{yh}  {\displaystyle \vh}_t + {\displaystyle \vb}^{y}), \label{eq:softmax}
\end{eqnarray}

\noindent where ${\displaystyle \mW}^{**}$ and ${\displaystyle \vb}^{*}$ are weight matrices and biases, respectively, and ${\displaystyle \vx}_t$ is the input vector at time $t$. The $REC$ module represents an arbitrary recurrent layer, with ${\displaystyle \sS}_{t-1}$ being a set of vectors describing its internal state at the previous time step. In the most general case, this module will simply be an LSTM cell, but we consider other possibilities as well, as described below.

The \textit{dual} architecture introduces an additional layer, with ReLU activation, which is fed with both the output of the embedding layer and the output of the recurrent module:

\begin{eqnarray}
    {\displaystyle \ve}_t &=& {\displaystyle \mW}^{ex} {\displaystyle \vx}_t \label{eq:embedding-dual} \\
    {\displaystyle \vh}_t &=& REC({\displaystyle \ve}_t, {\displaystyle \sS}_{t-1}) \label{eq:rec-dual}\\
    {\displaystyle \vd}_t &=& ReLU({\displaystyle \mW}^{de}{\displaystyle \ve}_t + {\displaystyle \mW}^{dh}{\displaystyle \vh}_t + {\displaystyle \vb}^{d})\\
    {\displaystyle \vy}_t &=& softmax({\displaystyle \mW}^{yd}  {\displaystyle \vd}_t + {\displaystyle \vb}^{y}). \label{eq:softmax-dual}
\end{eqnarray}

This way the network's input can reach the softmax layer following two different paths, through the recurrent layer and through the \textit{dual} connection. In the following we consider different forms for the recurrent module in equations \ref{eq:rec} and \ref{eq:rec-dual}.

\subsection{The LSTM module}
\label{subsec:LSTM}

In the simplest approach the recurrent module consists of an LSTM cell, where the internal state includes both the output and the memory, ${\displaystyle \sS}_{t} = \{{\displaystyle \vh}_{t}; {\displaystyle \vc}_{t}\}$, which are computed as follows:

\begin{eqnarray}
    {\displaystyle \vf}_{t} &=& \sigma({\displaystyle \mW}^{fe} {\displaystyle \ve}_{t} + {\displaystyle \mW}^{fh} {\displaystyle \vh}_{t-1} + {\displaystyle \vb}^{f}) \label{eq:lstm:1}\\
    {\displaystyle \vi}_{t} &=& \sigma({\displaystyle \mW}^{ie} {\displaystyle \ve}_{t} + {\displaystyle \mW}^{ih} {\displaystyle \vh}_{t-1} + {\displaystyle \vb}^{i}) \label{eq:lstm:2}\\
    {\displaystyle \vo}_{t} &=& \sigma({\displaystyle \mW}^{oe} {\displaystyle \ve}_{t} + {\displaystyle \mW}^{oh} {\displaystyle \vh}_{t-1} + {\displaystyle \vb}^{o}) \label{eq:lstm:3}\\
    {\displaystyle \vz}_{t} &=& tanh({\displaystyle \mW}^{ze} {\displaystyle \ve}_{t} + {\displaystyle \mW}^{zh} {\displaystyle \vh}_{t-1} + {\displaystyle \vb}^{z}) \label{eq:lstm:4}\\
    {\displaystyle \vc}_{t} &=& {\displaystyle \vf}_{t} \odot {\displaystyle \vc}_{t-1} + {\displaystyle \vi}_{t} \odot {\displaystyle \vz}_{t} \label{eq:lstm:5}\\
    {\displaystyle \vh}_{t} &=& {\displaystyle \vo}_{t} \odot tanh({\displaystyle \vc}_{t}),\label{eq:lstm:6}
\end{eqnarray}

\noindent where, as before, ${\displaystyle \mW}^{**}$ are weight matrices and ${\displaystyle \vb}^{*}$ are bias vectors. The $\odot$ operator denotes an element-wise product, and $\sigma$ is the logistic sigmoid function. For convenience, we summarize the joint effect of equations \ref{eq:lstm:1}-\ref{eq:lstm:6} as:

\begin{equation}
    {\displaystyle \vh}_t = LSTM({\displaystyle \ve}_t, \{{\displaystyle \vh}_{t-1}; {\displaystyle \vc}_{t-1}\}).
    \label{eq:module:1}
\end{equation}

In the literature it is quite common to stack several LSTM layers. Here we consider a double-layer LSTM, where the output ${\displaystyle \vh}_t$ of the recurrent module is obtained by the concatenated application of two LSTM layers:

\begin{eqnarray}
    {\displaystyle \vh}'_t &=& LSTM_{1}({\displaystyle \ve}_t, \{{\displaystyle \vh}'_{t-1}; {\displaystyle \vc}'_{t-1}\})\\
    {\displaystyle \vh}_t &=& LSTM_{2}({\displaystyle \vh}'_t, \{{\displaystyle \vh}_{t-1}; {\displaystyle \vc}_{t-1}\}).
\end{eqnarray}

We refer to this double LSTM module as $dLSTM$:

\begin{eqnarray}
    {\displaystyle \vh}_t &=& dLSTM({\displaystyle \ve}_t, \{{\displaystyle \vh}_{t-1}; {\displaystyle \vc}_{t-1}; {\displaystyle \vh}'_{t-1}; {\displaystyle \vc}'_{t-1}\}) \label{eq:module:2}\\
    &=& LSTM_{2}(LSTM_{1}({\displaystyle \ve}_t, \{{\displaystyle \vh}'_{t-1}; {\displaystyle \vc}'_{t-1}\}), \{{\displaystyle \vh}_{t-1}; {\displaystyle \vc}_{t-1}\}) .
\end{eqnarray}

\subsection{The Mogrifier-LSTM module}
\label{subsec:Mogrifier}

The Mogrifier-LSTM \cite{mogrifier} is one of the state-of-the-art variations of the standard LSTM architecture achieving the highest perplexity scores in language modeling tasks. It basically consists of a standard LSTM block, but the input ${\displaystyle \ve}_{t}$ and the hidden state ${\displaystyle \vh}_{t-1}$ are transformed before entering equations \ref{eq:lstm:1}-\ref{eq:lstm:6}. The mogrifier transformation involves several steps where ${\displaystyle \ve}_{t}$ and ${\displaystyle \vh}_{t-1}$ modulate each other:

\begin{eqnarray}
    \label{eq:x_mogrifier}
    {\displaystyle \ve}_{t}^{i} &=& 2 \sigma({\displaystyle \mQ}^{i}  {\displaystyle \vh}_{t-1}^{i-1}) \odot  {\displaystyle \ve}_{t}^{i-2}, \;\;\;\;\; \textrm{for odd} \; i \in \{1, 2, ..., r\} \\
    \label{eq:h_mogrifier}
     {\displaystyle \vh}_{t-1}^{i} &=& 2 \sigma({\displaystyle \mR}^{i}  {\displaystyle \ve}_{t}^{i-1} ) \odot  {\displaystyle \vh}_{t-1}^{i-2}, \;\;\;\;\; \textrm{for even} \; i \in \{1, 2, ..., r\}, 
\end{eqnarray}

\noindent where ${\displaystyle \mQ}^{i}$ and ${\displaystyle \mR}^{i}$ are weight matrices and we have ${\displaystyle \ve}_{t}^{-1} = {\displaystyle \ve}_{t}$ and ${\displaystyle \vh}_{t-1}^{0} = {\displaystyle \vh}_{t-1}$. The linear transformations ${\displaystyle \mQ}^{i}  {\displaystyle \vh}_{t-1}^{i-1}$ and ${\displaystyle \mR}^{i}  {\displaystyle \ve}_{t}^{i-1}$ can also include the addition of a bias vector, which has been omitted for the sake of clarity. The constant $r$ is a hyperparameter whose value defines the number of rounds of the transformation. We refer to this recurrent module, including the mogrifier transformation and the subsequent application of the LSTM layer, as:

\begin{equation}
    {\displaystyle \vh}_t = mLSTM({\displaystyle \ve}_t, \{{\displaystyle \vh}_{t-1}; {\displaystyle \vc}_{t-1}\}) = LSTM({\displaystyle \ve}_t^{*}, \{{\displaystyle \vh}_{t-1}^{*}; {\displaystyle \vc}_{t-1}\}), \label{eq:module:3}
\end{equation}

where ${\displaystyle \ve}_t^{*}$ and ${\displaystyle \vh}_{t-1}^{*}$ are the highest indexed ${\displaystyle \ve}_t^{i}$ and ${\displaystyle \vh}_{t-1}^{i}$ in equations \ref{eq:x_mogrifier} and \ref{eq:h_mogrifier}. Note that the choice $r = 0$ recovers the standard LSTM model. 

\cite{mogrifier} also used a double-layer LSTM enhanced with the mogrifier transformation. This strategy can be summarized as follows:

\begin{eqnarray}
    {\displaystyle \vh}_t &=& mdLSTM({\displaystyle \ve}_t, \{{\displaystyle \vh}_{t-1}; {\displaystyle \vc}_{t-1}; {\displaystyle \vh}'_{t-1}; {\displaystyle \vc}'_{t-1}\}) \label{eq:module:4}\\
    &=& mLSTM_{2}(mLSTM_{1}({\displaystyle \ve}_t, \{{\displaystyle \vh}'_{t-1}; {\displaystyle \vc}'_{t-1}\}), \{{\displaystyle \vh}_{t-1}; {\displaystyle \vc}_{t-1}\}) .
\end{eqnarray}

\section{Experiments}
\label{sec:experiments}

\subsection{Datasets}
\label{sec:datasets}

We perform experiments on two datasets: the Penn Treebank corpus \cite{ptb_corpus}, as  preprocessed by \cite{dyneval_Mikolov10}, and the WikiText-2 dataset \cite{merity_wikitext2}. In both cases, the data are used without any additional preprocessing.


The Penn Treebank (PTB) dataset has been widely used in the literature to experiment with language modeling. The standard data preprocessing is due to \cite{dyneval_Mikolov10}, and includes transformation of all letters to lower case, elimination of punctuation symbols, and replacement of all numbers with a special token. The vocabulary is limited to the 10,000 most frequent words. The data is split into a training set which contains almost 930,000 tokens, and  validation and test sets with around 80,000 words each. 


The WikiText-2 (WT2) dataset, introduced by \cite{merity_wikitext2}, is a more realistic benchmark for language modeling tasks. It consists of more than 2 million words extracted from Wikipedia articles. The training, validation and test sets contain around 2,125,000, 220,000, and 250,000 words, respectively. The vocabulary includes over 30,000 words, and the data retain capitalization, punctuation, and numbers.

\subsection{Experimental setup}
\label{sec:setup}

All the considered models follow one of the two architectures discussed in section \ref{sec:models}, either the Embedding-Recurrent-Softmax (ERS) architecture (equations \ref{eq:embedding}-\ref{eq:softmax}) or the \textit{dual} architecture (equations \ref{eq:embedding-dual}-\ref{eq:softmax-dual}). In either case, the recurrent module can be any of $LSTM$, $dLSTM$, or $mdLSTM$.
Weight tying \cite{inan2017tying,press_wt} is used to couple the weight matrices of the embedding and the output layers. This reduces the number of parameters and prevents the model from learning a one-to-one correspondence between the input and the output \cite{awd_lstm}. 

We run two different sets of experiments. First, we analyze the effect of the \textit{dual} connection by 
comparing the performances of the two architectures (ERS vs Dual), using each of the recurrent modules, on both the PTB and the WT2 datasets. In this setting the hyperparameters are tuned for the ERS architecture, and then transferred to the \textit{dual} case. Second, we search for the best hyperparameters for the \textit{dual} architecture using the $mdLSTM$ recurrence, and compare the perplexity score with current state-of-the-art values. All the experiments have been performed using the Keras library \cite{keras}. 

The networks are trained using the Nadam optimizer \cite{nadam}, a variation of Adam \cite{adam} where Nesterov momentum is applied. The number of training epochs is different for each experimental condition. On one hand, when the objective is to perform a pairwise comparison between \textit{dual} and non-dual architectures, we train the models for 100 epochs. On the other hand, when the goal is to compare the \textit{dual} network with state of the art approaches, we let the models run for 300 epochs. We use batch sizes of 32 and 128 for the PTB and the WT2 problems, respectively, and set the sequence length to 25 in all cases. The remaining hyperparameters are searched in the ranges described in tables \ref{tab:hyperparams} and \ref{tab:hyperparams_2} in appendix \ref{ap:hyperparams}. 

Finally, all the models are run twice, both with and without dynamic evaluation \cite{dyneval_Krause17}. Dynamic evaluation is a standard method commonly used to adapt the model parameters, learned during training, using also the validation data. This allows the networks to get adapted to the new evaluation conditions, which in general improves their performance. In order to keep the models as simple as possible, no additional modifications have been considered.

\section{Results}
\label{sec:results}

We first show the results of the comparative analysis ERS vs Dual, then we focus on the search of the optimal hyperparameters for the \textit{dual} architecture with the $mdLSTM$ recurrence.

\subsection{Dual vs non-dual architectures}

Table \ref{tab:model_vs_dual} displays the validation and test perplexity scores obtained for each of the experimental configurations on the PTB and the WT2 problems, both with and without dynamic evaluation. To facilitate the comparison, each pair of rows contain the results for one of the recurrent modules ($LSTM$, $dLSTM$ or $mdLSTM$) using the two architectures ERS and Dual, with the best values shown in bold. In each case, the hyperparameters are tuned for the standard ERS architecture and then used within the \textit{dual} networks without any additional adaptation. The exceptions are hyperparameters, such as the \textit{dual} dropout, which do not exist in the ERS configuration (those marked with an asterisk in tables \ref{tab:hyperparams} and \ref{tab:hyperparams_2} in appendix \ref{ap:hyperparams}).
To give a measure of the model complexity, table \ref{tab:model_vs_dual} contains also the approximate number of trainable parameters for each configuration. 

\begin{table}[t]
\caption{Validation and test word-level perplexity obtained for each of the experimental configurations on the PTB (top) and the WT2 (bottom) datasets.}
\label{tab:model_vs_dual}
\begin{center}
\begin{tabular}{llllll}
\multicolumn{6}{c}{\bf Penn Treebank Dataset} \\
&  & \multicolumn{2}{c}{\bf No Dyneval} & \multicolumn{2}{c}{\bf Dyneval} \\
\bf MODEL & \bf No. PARAMS & Val. & Test & Val. & Test \\
\hline \\
\textit{LSTM}             & 8.88 M & 67.37 & 64.91 & 62.31 & 61.17\\
\textit{Dual LSTM}        & 9.60 M & 61.22 & 59.39 & \bf 55.26 & \bf 54.69 \\
\hline \\
\textit{dLSTM}          & 13.62 M & 63.44 & 61.03 & 57.18 & 56.01 \\
\textit{Dual dLSTM}     & 13.94 M & 60.99 & 59.56 & \bf 56.11 & \bf 54.87 \\
\hline \\
\textit{mdLSTM}  & 21.43 M & 56.04 & 54.17 & 49.62 & 48.87 \\
\textit{Dual mdLSTM}        & 22.88 M & 54.16 & 52.16 & \bf 46.91 & \bf 46.09 \\
\hline \\
\multicolumn{6}{c}{\bf WikiText-2 Dataset} \\
&  & \multicolumn{2}{c}{\bf No Dyneval} & \multicolumn{2}{c}{\bf Dyneval} \\
\bf MODEL & \bf No. PARAMS & Val. & Test & Val. & Test \\
\hline \\
\textit{LSTM}             & 20.23 M & 92.84 & 88.28 & 74.98 & 69.42 \\
\textit{Dual LSTM}        & 20.95 M & 85.88 & 82.48 & \bf 61.94 & \bf 57.61 \\
\hline \\
\textit{dLSTM}          & 29.60 M & 78.65 & 75.60 & 63.26 & 59.42 \\
\textit{Dual dLSTM}     & 30.32 M & 77.01 & 73.90 & \bf 61.10 & \bf 57.10 \\
\hline \\
\textit{mdLSTM}  & 37.51 M & 72.05 & 69.06 & 57.42 & 53.93 \\
\textit{Dual mdLSTM}        & 38.95 M & 71.78 & 70.83 & \bf 53.48 & \bf 50.71 \\
\hline
\end{tabular}
\end{center}
\end{table}

As expected, dynamic evaluation improves the results regardless of the model or the dataset. The main observation, however, is that networks enhanced with the \textit{dual} connection display lower perplexity scores for almost all the training conditions on both the PTB and the WT2 datasets. 
The advantage of the Dual vs the ERS architecture is larger for less complex models, and narrows as the model complexity increases. Nevertheless, even for networks with $mdLSTM$ recurrence, the \textit{dual} architectures outperform their non-dual counterparts in more than 2 perplexity points on the test set, when dynamic evaluation is used.

Finally, it is worth noting that all the results presented correspond to our own implementation of the models, and that in most cases we are not including some of the several training or validation adaptations frequently used in the literature (such as AWD or MoS, for example). This can explain the difference with respect to the results reported by \cite{mogrifier} for the Mogrifier-LSTM model. We would expect a further improvement of the results were these additional mechanisms implemented.

\subsection{Dual Mogrifier fine tuning}

The second part of the experiments consists of searching for the best hyperparameters in the configuration that provided the smallest perplexity in the previous setup, that is the \textit{Dual mdLSTM} architecture. We carry out this experiment with the PTB problem. 
After an extensive search (see tables \ref{tab:hyperparams} and \ref{tab:hyperparams_2} in appendix \ref{ap:hyperparams}), the best performance is obtained with a model with 850 units in the embedding layer, 850 units in each of the mogrifier LSTM layers, and 850 units also in the \textit{dual} layer. The input, recurrent, internal, and output dropout rates are all set to 0.5, the \textit{dual} input and output dropout rates are set to 0.5 and 0.4, respectively, and the mogrifier dropout rate is set to 0.15. Both the embedding and the \textit{dual} L2 regularization parameters are set to $10^{-5}$. The mogrifier number of rounds is set to 4, and the rank to 100. All the remaining hyperparameters are set to 0. 

After the training phase, we continue with a fine tuning of some additional hyperparameters, using the validation data. First, we look for the best sequence length in the range $[5, 70]$, and then we fine-tune the softmax temperature in the range $[0.9, 1.3]$. When using dynamic evaluation, we also look for the best gradient clipping value (in the range $[0.0, 1.0]$) and, following \cite{mogrifier}, we repeat the whole procedure with the $\beta_1$ parameter of the Nadam optimizer set to $0$, which resembles the RMSProp optimizer without momentum. 
The results are shown in table \ref{tab:state-of-the-art}, together with the top perplexity scores reported in the literature for the same problem.

\begin{table}[t]
\caption{Best validation and test word-level perplexity scores reported in the literature for the Penn Treebank dataset, with and without dynamic evaluation. Missing values in the last two columns correspond to works where the dynamic evaluation approach was not considered. The last row in the table displays the results obtained with our \textit{Dual mdLSTM} network.}
\label{tab:state-of-the-art}
\begin{center}
\begin{tabular}{lrcccc}
&  & \multicolumn{2}{c}{\bf No Dyneval} & \multicolumn{2}{c}{\bf Dyneval} \\
\bf MODEL & & Val. & Test & Val. & Test \\
\hline \\
\textit{AWD-LSTM} \cite{awd_lstm}                           & 24 M  & 60.00 & 57.30 & - & - \\
\textit{AWD-LSTM-DOC} \cite{takase-etal-2018-direct}        & 23 M  & 54.12 & 52.38 & - & - \\
\textit{AWD-LSTM} \cite{dyneval_Krause17}                   & 24 M  & 59.80 & 57.70 & 51.60 & 51.10 \\
\textit{AWD-LSTM+PDR} \cite{brahma-2019-improved}          & 24 M  & 57.90 & 55.60 & 50.10 & 49.30 \\
\textit{AWD-LSTM+MoS} \cite{awd_lstm_mos}                  & 22 M  & 56.54 & 54.44 & 48.33 & 47.69 \\
\textit{AWD-LSTM+MoS+PDR} \cite{brahma-2019-improved}     & 22 M  & 56.20 & 53.80 & 48.00 & 47.30 \\
\textit{AWD-LSTM-DOC x5} \cite{takase-etal-2018-direct}     & 185 M & 48.63 & 47.17 & - & - \\
\textit{AWD-LSTM+MoS+FRAGE} \cite{frage}                  & 24 M  & 55.52 & 53.51 & 47.38 & 46.54 \\
\textit{AWD-LSTM+MoS+Adv} \cite{pmlr-v97-wang19f}         & 22 M  & 54.98 & 52.87 & 47.15 & 46.52 \\
\textit{AWD-LSTM+MoS+Adv+PS} \cite{pmlr-v97-wang19f}     & 22 M  & 54.10 & 52.20 & 46.63 & 46.01 \\
\textit{Mogrifier-LSTM} \cite{mogrifier}                    & 24 M  & 51.40 & 50.10 & 44.90 & 44.80 \\
\bf \textit{Dual mdLSTM} - ours                       & 23 M & 52.87 & 51.19 & 45.13 & \bf 44.61 \\
\hline
\end{tabular}
\end{center}
\end{table}

The state-of-the-art is dominated by several variations of the AWD-LSTM network \cite{awd_lstm}, the most common being the inclusion of a Mixture of Softmaxes (MoS) \cite{awd_lstm_mos}. Other add-ons include Direct Output Connection (DOC) \cite{takase-etal-2018-direct}, which is a generalization of MoS, Frequency Agnostic word Embedding (FRAGE) \cite{frage}, Past Decode Regularization (PDR) \cite{brahma-2019-improved}, or Partial Shuffling (PS) with Adversarial Training (Adv) \cite{pmlr-v97-wang19f}. The mogrifier-LSTM described in section \ref{subsec:Mogrifier} combines many of these ideas with a mutual gating between the input and the hidden state vectors to obtain the best results reported in the literature for the PTB problem, among those obtained by networks that do not use additional data during the training phase. Compared with all these models, our current approach leads the ranking with a perplexity score of $44.61$, even though most of the aforementioned tricks have not been considered.

\section{Discussion}
\label{sec:conclusion}

In this work, we have presented a new network design for the Language Modeling task based on the \textit{dual} network proposed by \cite{Oliva21_DualRNN}. This network adds a direct connection between the input and the output, skipping the recurrent module, and can be adapted to any of the traditional Embedding-Recurrent-Softmax (ERS) models, opening the way to new approaches for this task. We have based our work on the Penn Treebank \cite{dyneval_Mikolov10} and the WikiText-2 \cite{merity_wikitext2} datasets, comparing the ERS approach and its \textit{dual} alternative. Regardless of the configuration, the \textit{dual} version performs always better, even though it faces a slight disadvantage, since most of the hyperparameters are tuned using the ERS model. We can expect a much better performance if the complete set of hyperparameters is properly tuned for the \textit{dual} network.

This is in fact the case for the second experiment, where a \textit{Dual mdLSTM}, which includes a simplified version of the mogrifier LSTM \cite{mogrifier} within a \textit{dual} architecture, is fine tuned for the Penn Treebank dataset. After a thorough search of the hyperparameters space, we have found a network configuration that establishes a new state-of-the-art score for this problem. Interestingly, this new record has been obtained in spite of leaving aside many of the standard features used in most state-of-the-art approaches, such as the Averaged SGD Weight-Drop (\textit{AWD}) \cite{awd_lstm} or the Mixture of Softmaxes (\textit{MoS}) \cite{awd_lstm_mos}. The incorporation of these features into the \textit{dual} architecture can be expected to further increase the model performance. 

The \textit{dual} architecture was firstly proposed as an alternative that reduces the computational load on the recurrent layer, letting it concentrate on modeling the temporal dependencies only. From a more abstract point of view, it has been argued that the dual architecture can be understood as a sort of Mealy machine, where the output explicitly depends on both the hidden state and the input \cite{Oliva21_DualRNN}. Our results show that this explicit dependence on the input can indeed lead to better performance on language modeling tasks. This emphasizes the importance of the current input in RNN models.

Finally, although the new approach has not been tested with large-scale language corpora, we expect that our results scale well to larger datasets. Work in progress contemplates this extension. The \textit{dual} architecture also needs further research concerning the deepness of the specific variations of Language Modeling and other families of problems not necessarily related to Natural Language Processing. This work opens a new line of research to be considered when processing any sequence or time series. The utility of this approach in more general problems will be addressed in future work.



\bibliography{iclr2022_conference}
\bibliographystyle{abbrv}

\newpage

\appendix
\section{Appendix}
\label{ap:hyperparams}

\begin{table}[bth!]
\caption{List of all the hyperparameters and the search range associated with each of them. Those marked with an asterisk $(^*)$ refer to the \textit{dual} architectures only.}
\label{tab:hyperparams}
\begin{center}
\begin{tabular}{lll}
\bf Name & \bf Description & \bf Values \\
\hline \\
\textit{Num epochs} & Number of training epochs. & \{100, 300\} \\
\textit{Learning rate} & Learning rate. & [10$^{-6}$, 10$^{-3}$] \\
\textit{Batch size} & Batch size. & \{32, 128\} \\
\textit{Seq len} & Sequence length. & \{10, 25, 50\} \\
\hline \\
\textit{Embedding units} & Size of the embedding layer. & \{400, 850\} \\
\textit{Recurrent units} & Size of the recurrent layers. & \{400, 850, 1150\} \\
\textit{LSTM layers} & Number of recurrent layers. & \{1, 2, 3\} \\
\textit{Dual units}$^*$ & Size of the \textit{dual} layer. & \{400, 850\} \\
\hline \\
\textit{Embedding L2reg} & L2 regularization applied to & \{0, $10^{-6}$, $10^{-5}$\} \\
& the Embedding and output & \\
& layers. & \\
\textit{Rec. input L2reg} & L2 regularization applied to & \{0, $10^{-6}$, $10^{-5}$\} \\
& the input weights of the  & \\
& recurrent layer. & \\
\textit{Rec. L2reg} & L2 regularization applied to & \{0, $10^{-6}$, $10^{-5}$\} \\
& the recurrent weights of the & \\
& recurrent layer. & \\
\textit{Activation L2reg} & L2 regularization applied to & \{0, $10^{-6}$, $10^{-5}$\} \\
& the recurrent layers output. & \\
\textit{Dual L2reg}$^*$ & L2 regularization applied to & \{0, $10^{-6}$, $10^{-5}$\} \\
& \textit{dual} layer. & \\
\hline \\
\end{tabular}
\end{center}
\end{table}

\begin{table}[bth!]
\caption{List of all the hyperparameters and the search range associated with each of them. Those marked with an asterisk $(^*)$ refer to the \textit{dual} architectures only.}
\label{tab:hyperparams_2}
\begin{center}
\begin{tabular}{lll}
\bf Name & \bf Description & \bf Values \\
\hline \\
\textit{Rec. input Drop.} & Dropout before the first & [0.0, 0.5] \\
& recurrent layer. & \\
\textit{Rec. Drop.} & Dropout for the linear & [0.0, 0.5] \\
& transformation of the recurrent & \\
& state. & \\
\textit{Rec. internal Drop.} & Dropout between the recurrent & [0.0, 0.5] \\
& layers. & \\
\textit{Rec. output Drop.} & Dropout after the last recurrent & [0.0, 0.5] \\
& layer. & \\
\textit{Dual input Drop.}$^*$ & Dropout before the \textit{dual} layer. & [0.0, 0.5] \\
\textit{Dual output Drop.}$^*$ & Dropout after the \textit{dual} layer. & [0.0, 0.5] \\
\hline \\
\textit{Mogrifier deep} & Mogrifier rounds. & \{0, 2, 3, 4, 5, 6\} \\
\textit{Mogrifier L2reg} & L2 regularization applied to & \{0, $10^{-6}$, $10^{-5}$\} \\
& Mogrifier weights. & \\
\textit{Mogrifier rank} & Weight factorization. ${\displaystyle \mQ}^{i} = {\displaystyle \mQ}^{i}_l{\displaystyle \mQ}^{i}_r$  & \{0, 50, 100, 200\} \\
& with ${\displaystyle \mQ}^{i}_l\in \displaystyle \R^{m\times k}, {\displaystyle \mQ}^{i}_r\in \displaystyle \R^{k\times n}$. & \\
\textit{Mogrifier Drop.} & Dropout between the Mogrifier & [0.0, 0.2] \\
& weights. & \\
\hline \\
\textit{Learning rate eval} & Learning rate when Dynamic & [10$^{-6}$, 10$^{-3}$] \\
& evaluation. & \\
\textit{Seq len eval} & Sequence length when Dynamic & [5, 50] \\
& evaluation. & \\
\textit{Clipnorm eval} & Gradients clipping to a maximum & [0.0, 1.0] \\
& norm. & \\
\hline
\end{tabular}
\end{center}
\end{table}

\end{document}

%% file: math_commands.tex
\usepackage{amsmath,amsfonts,bm}

\def\eqref#1{equation~\ref{#1}}

\def\1{\bm{1}}

\def\vb{{\bm{b}}}
\def\vc{{\bm{c}}}
\def\vd{{\bm{d}}}
\def\ve{{\bm{e}}}
\def\vf{{\bm{f}}}

\def\vh{{\bm{h}}}
\def\vi{{\bm{i}}}

\def\vo{{\bm{o}}}

\def\vx{{\bm{x}}}
\def\vy{{\bm{y}}}
\def\vz{{\bm{z}}}

\def\mQ{{\bm{Q}}}
\def\mR{{\bm{R}}}

\def\mW{{\bm{W}}}

\DeclareMathAlphabet{\mathsfit}{\encodingdefault}{\sfdefault}{m}{sl}
\SetMathAlphabet{\mathsfit}{bold}{\encodingdefault}{\sfdefault}{bx}{n}

\def\sS{{\mathbb{S}}}

\newcommand{\R}{\mathbb{R}}

